\documentclass[10pt,twocolumn,letterpaper]{article}

\usepackage[pagenumbers]{cvpr} 

\definecolor{cvprblue}{rgb}{0.21,0.49,0.74}
\usepackage[pagebackref,breaklinks,colorlinks,allcolors=cvprblue]{hyperref}
\usepackage{multirow}

\def\paperID{3146} 
\def\confName{CVPR}
\def\confYear{2026}

\title{MICON-Bench: Benchmarking and Enhancing Multi-Image Context Image Generation in Unified Multimodal Models}

\author{Mingrui Wu$^{12}$\thanks{Equal Contribution}, Hang Liu$^{1*}$, Jiayi Ji$^1$, Xiaoshuai Sun$^1$, Rongrong Ji$^1$ \\
  $1$ Key Laboratory of Multimedia Trusted Perception and Efficient Computing, \\ Ministry of Education of China, Xiamen University, 361005, P.R. China.\\
  $2$ Zhongguancun Academy, Beijing, China.100094. \\
 } 

\begin{document}
\maketitle
\begin{abstract}
Recent advancements in Unified Multimodal Models (UMMs) have enabled remarkable image understanding and generation capabilities. However, while models like Gemini-2.5-Flash-Image show emerging abilities to reason over multiple related images, existing benchmarks rarely address the challenges of multi-image context generation, focusing mainly on text-to-image or single-image editing tasks. In this work, we introduce \textbf{MICON-Bench}, a comprehensive benchmark covering six tasks that evaluate cross-image composition, contextual reasoning, and identity preservation. We further propose an MLLM-driven Evaluation-by-Checkpoint framework for automatic verification of semantic and visual consistency, where multimodal large language model (MLLM) serves as a verifier. Additionally, we present \textbf{Dynamic Attention Rebalancing (DAR)}, a training-free, plug-and-play mechanism that dynamically adjusts attention during inference to enhance coherence and reduce hallucinations. Extensive experiments on various state-of-the-art open-source models demonstrate both the rigor of MICON-Bench in exposing multi-image reasoning challenges and the efficacy of DAR in improving generation quality and cross-image coherence. Github:\url{https://github.com/Angusliuuu/MICON-Bench}.
\end{abstract}

\section{Introduction}

The rapid evolution of Unified Multimodal Models (UMMs)~\cite{google2025gemini,hurst2024gpt,deng2025emerging,wu2025omnigen2,xie2025show,chen2025blip3,wang2024emu3,zhou2024transfusion,dong2023dreamllm,ma2025janusflow,wu2024next,team2024chameleon} has significantly advanced visual understanding and image synthesis. Recent native multimodal models, such as Nano-Banana~\cite{google2025gemini}, demonstrate the ability to reason over multiple images and textual instructions simultaneously, generating contextually coherent visual outputs. This progress signals a shift from single-image generation toward multi-image context generation, where the goal is to integrate and reason across multiple related images to produce consistent and semantically aligned outputs.

Despite these advances, the capability of current generative models to handle multi-image context remains underexplored and poorly quantified. Existing evaluations~\cite{ghosh2023geneval,huang2023t2i,sheynin2024emu,ye2025imgeditunifiedimageediting,liu2025step1xeditpracticalframeworkgeneral} primarily focus on text-to-image generation or single-image editing tasks, emphasizing fidelity and text alignment. However, multi-image generation introduces new challenges, such as cross-image consistency, spatial-temporal coherence, and reasoning across complex visual relationships (e.g., maintaining object identity and spatial continuity across multiple reference images). Without a dedicated benchmark, it is difficult to diagnose failure modes or compare methods fairly in this emerging setting.

To address this gap, we introduce MICON-Bench — a comprehensive benchmark specifically designed for evaluating multi-image context generation. MICON-Bench consists of six tasks that cover both simple and complex scenarios, ranging from basic object composition and attribute transfer to high-level visual storytelling generation. Each task is designed to probe distinct dimensions of multi-image reasoning, such as compositional understanding, reference consistency, and contextual inference.
Unlike prior benchmarks, MICON-Bench adopts an Evaluation-by-Checkpoint paradigm to ensure objective and scalable assessment. For each case, we define verifiable checkpoints representing key visual relationships or contextual dependencies, and automatically evaluate generated images via an MLLM-based verifier. Specifically, the multimodal large language model (MLLM)~\cite{Qwen2.5-VL,wang2025internvl3} judges whether each checkpoint condition is satisfied by the generated image, producing binary outcomes that are averaged into the final task score.


\begin{figure*}[t]
  \centering
  \includegraphics[width=0.96\textwidth]{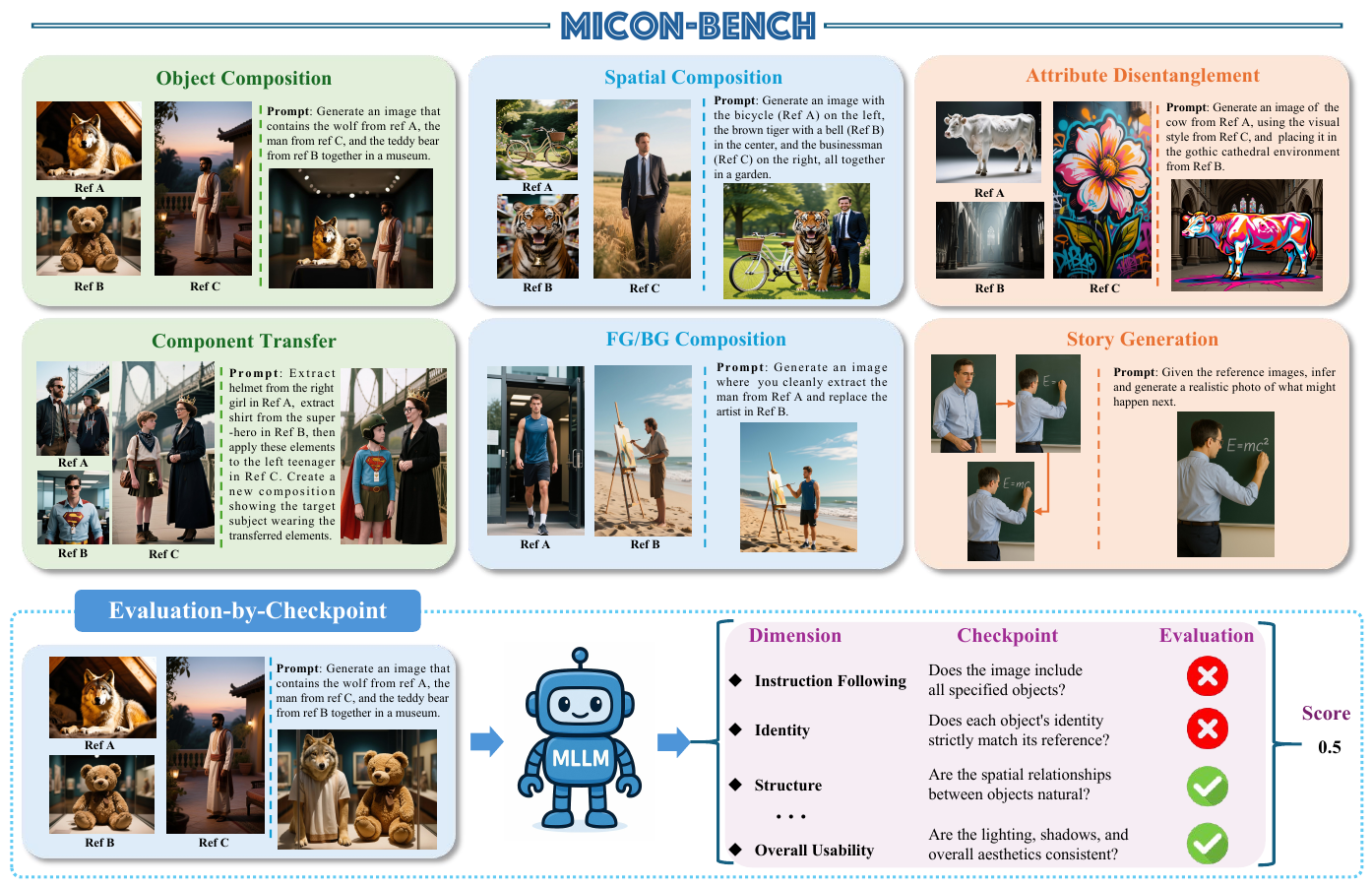}
  \caption{Overview of MICON-Bench and Evaluation Pipeline.
MICON-Bench is a comprehensive benchmark designed to evaluate multi-image context generation across six diverse tasks: Object Composition, Spatial Composition, Attribute Disentanglement, Component Transfer, FG/BG Composition, and Story Generation.
Each task provides multiple reference images and a compositional prompt requiring the model to integrate, reason, or transfer information across images.
To assess model performance, we introduce an Evaluation-by-Checkpoint framework driven by an MLLM, which automatically evaluates generated results along key dimensions, producing an interpretable composite score.}
  \label{fig:intro}
\end{figure*}

While MICON-Bench provides a standardized platform for evaluation, our analysis reveals that even state-of-the-art UMMs, including open-source multimodal generators~\cite{deng2025emerging,wu2025omnigen2}, struggle with cross-image consistency. In particular, UMMs tend to distribute attention uniformly across reference images, often focusing on irrelevant regions, which leads to visual inconsistencies or hallucinations in the synthesized results.

To mitigate these issues, we propose Dynamic Attention Rebalancing (DAR), a training-free, plug-and-play mechanism that adaptively reweights attention toward semantically relevant regions during inference. DAR dynamically identifies over- and under-attended areas in reference images based on attention maps and rebalances them via attention scaling. This enables UMMs to better preserve identity, spatial relations, and contextual coherence across multiple generated frames or references, without any additional training or fine-tuning.

In summary, our contributions are threefold:
\begin{itemize}
\item We introduce MICON-Bench, a comprehensive benchmark suite covering six diverse multi-image context generation tasks, supporting rigorous and scalable evaluation via an MLLM-driven Evaluation-by-Checkpoint verification framework.
\item We propose Dynamic Attention Rebalancing (DAR), a novel, training-free technique that effectively improves attention allocation in UMMs, significantly enhancing generation quality, identity preservation, and attribute coherence across multiple reference images.
\item We conduct extensive evaluations on multiple state-of-the-art image generation models, revealing significant challenges in multi-image reasoning and consistency.
\end{itemize}

\section{Related Work}
\label{sec:relate}

\subsection{Image Generation Benchmarks}
The evaluation of text-to-image generation has been primarily guided by benchmarks~\cite{chang2025oneig,chen2025multiref,ghosh2023geneval,hua2025mmig,liu2025step1xeditpracticalframeworkgeneral,ye2025imgeditunifiedimageediting,sheynin2024emu,li2025gir,ye2025echo,pan2025ice,liang2025idea,wang2025genspace,huang2023t2i}, which provide large-scale datasets with textual descriptions to evaluate the realism and semantic coherence of generated images. 
For evaluating text-visual consistency in the text-to-image generation, Geneval~\cite{ghosh2023geneval} and T2ICompBench~\cite{huang2023t2i} are widely used. They enable detailed assessment of how well generated images align with their text prompts. For single-image editing, ImgEdit-Bench~\cite{ye2025imgeditunifiedimageediting}, EMU Edit~\cite{sheynin2024emu}, and GEdit-Bench~\cite{liu2025step1xeditpracticalframeworkgeneral} evaluate the quality and faithfulness if text- or image-conditioned edits. Besides these single-image benchmarks, OmniContext~\cite{wu2025omnigen2} introduces a multi-image benchmark for image generation. However, its tasks mainly involve relatively simple subject-centric compositions, and the evaluation focuses on subject consistency across images. Overall, systematic evaluation of multi-image context generation across diverse tasks remains underexplored. To complement existing benchmarks, our MICON-Bench introduces a new suite of tasks that evaluate models' ability to synthesize images from multiple related inputs, providing a more realistic testbed for next-generation generative models.

\subsection{Image Generation Models}
Text-to-image generation models~\cite{rombach2022high,zhang2025context,kumari2025generating,yang2024cross} have advanced significantly with the introduction of models like Stable Diffusion~\cite{rombach2022high}, which uses latent diffusion to generate high-quality images from textual descriptions. These models~\cite{flux2024,esser2024scaling,podell2023sdxl,chen2024pixart,saharia2022photorealistic,liu2024playground,wu2025tradiffusion,wu2025reprompt} excel at generating detailed images from single text prompts. However, they are limited to single-image generation and do not address multi-image context or consistency.
Recently, native Unified Multimodal Models~\cite{deng2025emerging,wu2025omnigen2,hurst2024gpt,google2025gemini} have made strides in multi-image context generation, but challenges remain in ensuring coherence and consistency across multiple images.


\section{MICON-Bench}
To assess the multimodal context image generation capabilities of unified multimodal models, we propose MICON-Bench, which focuses on their ability to process multi-image inputs and generate outputs that are contextually consistent and aligned with the given instructions.

\begin{figure*}[t]
  \centering
  \includegraphics[width=0.96\textwidth]{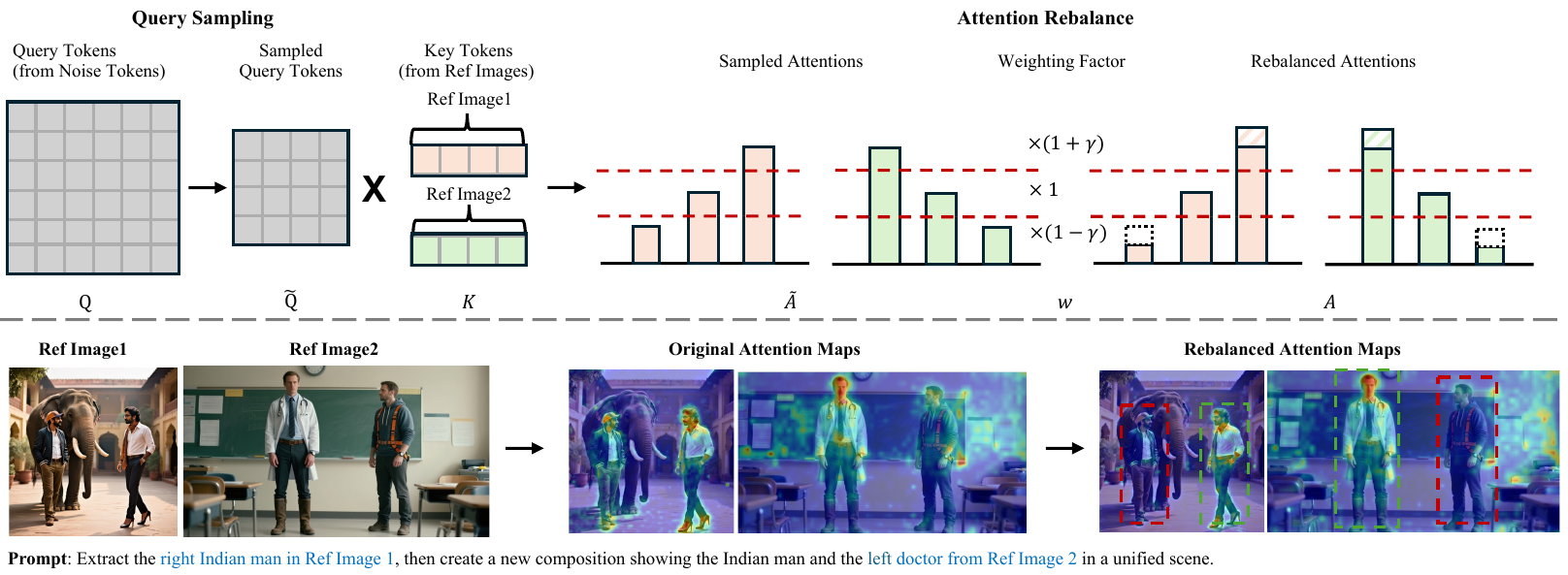}
  \caption{
Overview of the proposed \textbf{Dynamic Attention Rebalancing (DAR)} mechanism. 
Given multiple reference images, DAR first samples query tokens and computes attention maps between sampled queries and reference key tokens. 
It then applies a dynamic weighting factor to rebalance attention responses, reinforcing relevant reference regions (\textcolor{green}{green boxes}) while suppressing distractions (\textcolor{red}{red boxes}). 
The example at the bottom shows how DAR helps compose a new scene by integrating the right Indian man from Ref Image 1 and the left doctor from Ref Image 2, achieving coherent and faithful visual synthesis.
}
  \label{fig:method}
\end{figure*}

\subsection{Task Description}
As shown in Figure~\ref{fig:intro},
the MICON-Bench consists of 6 tasks—5 compositional tasks (object composition, spatial composition, attribute disentanglement, component transfer, and FG/BG composition) and 1 complex task (story generation, which requires generating images that follow causal reasoning). 
The final benchmark consists of 1,043 cases and 2,518 images. 

\vspace{-0.5em}
\paragraph{Object Composition.}
The task involves generating an image by combining a single subject (person, animal, or object) with a background scene (e.g., indoor or outdoor). The goal is to assess the model’s ability to create realistic object compositions in different environments. The task is constructed by randomly sampling an object from a predefined object pool  and a scene from a scene pool. These elements are combined to generate reference images using \texttt{Qwen-Image}~\cite{wu2025qwen}. The task includes 118 cases with two reference images and 82 cases with three reference images.

\vspace{-0.5em}
\paragraph{Spatial Composition.} 
Building upon object composition, this task introduces spatial constraints where multiple objects must be arranged according to specific geometric relations (e.g., front, left, right). It challenges the model to generate images with spatial consistency, ensuring that objects are positioned correctly. Reference images are generated by adding spatial constraints to the object composition and processed by \texttt{Qwen-Image}. There are 200 cases, with 102 using two reference images and 98 using three reference images.

\vspace{-0.5em}
\paragraph{Attribute Disentanglement.}
This task requires generating an image that combines a subject, style, and background from three separate reference images. The goal is to test the model’s ability to decouple visual attributes and recombine them coherently. It contains 100 cases, each using three reference images.
The first image provides the subject, the second provides the style reference, and the third provides the background environment. These reference images are generated by \texttt{Qwen-Image}.

\vspace{-0.5em}
\paragraph{Component Transfer.} 
This task involves transferring specific elements (such as accessories or parts of a subject) between different reference images and applying them to a target subject. The objective is to test the model’s capability to manipulate and recombine objects while maintaining naturalness and consistency. The reference images generally contain multiple subjects, accessories, and backgrounds, created with \texttt{Qwen-Image}. It consists of 240 cases, with 119 using two reference images and 121 using three reference images.

\vspace{-0.5em}
\paragraph{FG/BG Composition.} 
In this task, the model is asked to extract the foreground from one reference image and combine it with the background of another. The goal is to assess the model's ability to blend foreground and background elements seamlessly. We use \texttt{Qwen-Image} to generate reference images, including one with the foreground subject and another with the background scene. There are 200 cases, all using two reference images.

\vspace{-0.5em}
\paragraph{Story Generation.} 
This complex task involves generating a realistic image that infers what might happen next in a story, given a set of reference images. It tests the model’s ability to perform causal reasoning and predict narrative continuations. Story texts are LLM-generated and human-filtered, and then \texttt{GPT-4o-Image} and \texttt{Nano-Banana} generate four-panel comic-style reference image sets reflecting physical laws, commonsense, and causal narrative reasoning. Images are enhanced with \texttt{DiffBIR} for clarity and undergo human filtering for accuracy and consistency. The dataset consists of 103 cases, with 72 using two reference images and 31 using three.



\subsection{Evaluation Process}
The evaluation process for the benchmark is based on seven key dimensions that assess various aspects of image generation tasks. These dimensions include Instruction Following, which measures how accurately the model follows the original instructions; Identity, assessing the consistency of key objects or characters; Structure, evaluating spatial relationships and proportions; Cross-Reference Consistency, ensuring no contradictions across multiple reference images; Causality, which examines the logical flow of story events; Text Grounding, checking if the text in the image is accurate and clear; and Overall Usability, which considers the overall visual quality and usability of the generated images. 
Based on these seven dimensions, we apply MLLM to generate verifiable checkpoints for each case, with the specific prompt provided in the appendix.

We evaluate the score of each task by feeding the input, generated image, and corresponding checkpoints into MLLM. Each checkpoint is classified as either pass or fail, and the final score is obtained by averaging the scores across all checkpoints. For complex task, which involves evaluating the model's reasoning abilities, we not only set the checkpoints but also incorporate an evaluation based on a predefined answer set. Details are shown in the Appendix.

\section{Method}
Currently, state-of-the-art UMMs for multi-image context generation leverage attention mechanisms to control the model's focus on various reference images. However, as illustrated in Figure~\ref{fig:vis2}, we observe that these UMMs occasionally attend indiscriminately to irrelevant regions within the reference images, resulting in hallucinated content in the generated outputs.

\paragraph{Dynamic Attention Rebalancing.}
To mitigate the above issue, we propose a plug-and-play Dynamic Attention Rebalancing (DAR) mechanism (Figure~\ref{fig:method}) that adaptively modulates the model’s attention over reference image regions during inference. DAR dynamically amplifies attention on regions pertinent to the generation task while suppressing attention to irrelevant areas, thereby achieving a balanced and focused attention distribution.

We first analyze the attention maps between query tokens (i.e., noise tokens) and reference image tokens to identify which regions require enhancement or suppression. A straightforward computation of attention between all query and reference tokens is computationally prohibitive. To alleviate this, we uniformly sample a subset of \(m \ll L_q\) query tokens, drastically reducing the computation while preserving representative attention statistics.
Formally, let \(Q \in \mathbb{R}^{L_q \times d}\) denote the query matrix, where \(L_q\) is the number of query tokens and \(d\) is the dimensionality per attention head. The sampled query subset \(\tilde{Q} \in \mathbb{R}^{m \times d}\) is selected uniformly via indices:
$\left\{\left\lfloor i \cdot \frac{L_q - 1}{m - 1} \right\rfloor \right\}_{i=0}^{m-1}$.


This subset of query tokens \(\tilde{Q}\) is then used to compute attention scores with the reference image tokens. Let \(K_{\text{ref}} \in \mathbb{R}^{L_{\text{ref}} \times H \times d}\) denote the key matrix for the reference image, where \(L_{\text{ref}}\) is the number of reference image tokens, \(H\) is the number of attention heads, and \(d\) is the dimensionality of the key vectors. 
The attention map \(\tilde{A}\) is computed as:
\begin{equation}
    \tilde{A} = \text{softmax} \left( \frac{\tilde{Q} K_{ref}^{\top}}{\sqrt{d}} \right),
\end{equation}
where the attention weight \(\tilde{A}_{i,h,k}\) denotes the attention between the \(i\)-th sampled query, the \(h\)-th attention head, and the \(k\)-th reference image token.

The total attention score \(r_k\) for the \(k\)-th reference token is computed by summing the attention weights across all attention heads:

\begin{equation}
    r_k = \sum_{i=1}^{m} \sum_{h=1}^{H} \tilde{A}_{i,h,k}.
\end{equation}

We then use min-max normalization to obtain the normalized attention score \(\hat{r}_k\):

\begin{equation}
    \hat{r}_k = \frac{r_k - \min_j r_j}{\max_j r_j - \min_j r_j}.
\end{equation}

To dynamically adjust the model's attention to the most relevant areas of the reference image, we define two thresholds, \(\tau_{\mathrm{high}}\) and \(\tau_{\mathrm{low}}\), to partition reference tokens into highly relevant, irrelevant, and neutral regions based on their normalized attention scores \(\hat{r}_k\):
\begin{equation}
    w_k = \begin{cases}
    1 + \gamma, & \text{if } \hat{r}_k \geq \tau_{\mathrm{high}}, \\
    1 - \gamma, & \text{if } \hat{r}_k \leq \tau_{\mathrm{low}}, \\
    1, & \text{otherwise},
    \end{cases}
\end{equation}
where \(\gamma \in [0,1]\) controls the magnitude of attention adjustment.
This weighting factor \(w_k\) dynamically modulates the influence of each reference token in the final attention computation.
The adjusted attention map is then calculated using the full query set \(Q\) as:
\begin{equation}
    A = \mathrm{softmax} \left( \frac{Q (w \odot K_{\mathrm{ref}})^{\top}}{\sqrt{d}} \right),
\end{equation}
where the weighting vector \(w = [w_1, w_2, \ldots, w_{L_{\mathrm{ref}}}]\) is broadcast appropriately over the reference key matrix, and \(\odot\) denotes element-wise multiplication.

This attention adjustment improves focus on relevant regions and suppresses irrelevant ones, enhancing generation quality with minimal additional computational cost.

\section{Experiments}
\subsection{Experimental Setup}
We conduct comparative experiments on two state-of-the-art autoregressive text-to-image generation models: BAGEL~\cite{deng2025emerging} and OmniGen2~\cite{wu2025omnigen2}. Both models represent strong baselines that utilize multi-image context for compositional image generation. We leverage the powerful multi-modal large language model Qwen3-VL-32B-Instruct~\footnote{https://huggingface.co/Qwen/Qwen3-VL-32B-Instruct.} to perform zero-shot multi-dimensional scoring. All experiments are conducted on 4 NVIDIA A800(40G) GPUs. Unless otherwise specified, we set the attention weight modulation factor \(\gamma=0.15\), sample \(m=64\) query tokens, $\tau_{\mathrm{high}}=0.7$ and $\tau_{\mathrm{low}}=0.3$.

\begin{figure*}[t]
  \centering
  \includegraphics[width=0.88\textwidth]{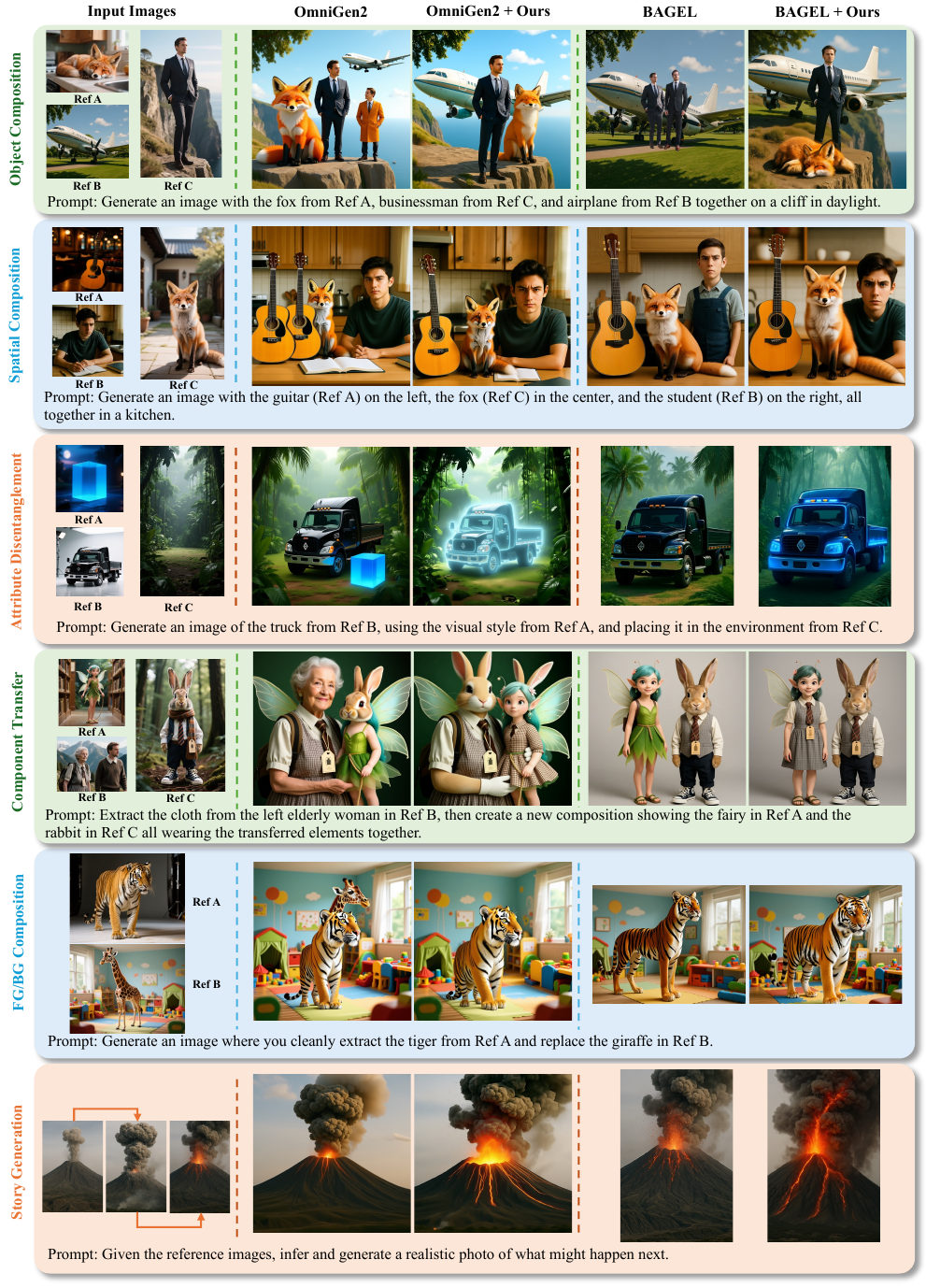}
  \caption{
Visualization examples of our method vs. baseline on MICON-Bench.
}
  \label{fig:vis}
\end{figure*}

\subsection{Main Results}
\paragraph{Performance Comparison on MICON-Bench.}
We evaluate several multi-image generation models on MICON-Bench, including proprietary UMMs (Nano-Banana, GPT-Image), diffusion-based models (UNO, Qwen-Image-Edit, DreamOmni2), and open-source UMMs (OmniGen2, BAGEL), where we test our proposed DAR method. Results are shown in Table~\ref{tab:performance_comparison}. Nano-Banana and GPT-Image lead overall, especially in Object and Spatial tasks, showcasing strong semantic and spatial reasoning. Diffusion-based models like UNO underperform, highlighting challenges in complex multimodal reasoning. Our DAR consistently improves OmniGen2 and BAGEL, with notable gains in Component and FG/BG tasks for OmniGen2, and boosted Story and FG/BG scores for BAGEL. These results confirm that DAR effectively enhances object perception, spatial understanding, and attribute reasoning in multi-image context generation with UMMs.

\begin{table*}[t]
\centering
\caption{Performance comparison of different models on our MICON-Bench. }
\begin{tabular}{lccccccc}
\toprule
\textbf{Model} & \textbf{Object} & \textbf{Spatial} & \textbf{Attribute} & \textbf{Component} & \textbf{FG/BG} & \textbf{Story} & \textbf{Avg. Score} \\
\midrule

Nano-Banana~\cite{google2025gemini}         & 95.60 & 93.79 & 92.13 & 84.23 & 83.13 & 82.84 & 89.25 \\
GPT-Image~\cite{hurst2024gpt}           & 96.45 & 94.41 & 93.39 & 87.69 & 90.15 & 85.99 & 91.51 \\
UNO~\cite{wu2025less}                 & 58.40 & 66.68 & 65.28 & 28.84 & 20.96 & 39.08 & 44.76 \\
DreamOmni2~\cite{xia2025dreamomni2}          & 88.24 & 84.76 & 85.28 & 59.64 & 76.16 & 59.58 & 75.56 \\
Qwen-Image-Edit-2507~\cite{wu2025qwen} & 96.52 & 88.80 & 78.04 & 42.68 & 72.08 & 63.81 & 72.96 \\

\midrule
BAGEL~\cite{deng2025emerging}               & 87.64 & 89.96 & 89.84 & 52.40 & 64.64 & 65.09 & 73.55 \\
BAGEL + \textbf{DAR}        & \textbf{88.04} & \textbf{91.88} & \textbf{90.76} & \textbf{56.06} & \textbf{71.24} & \textbf{66.34} & \textbf{76.31} \\
\midrule
OmniGen2~\cite{wu2025omnigen2}            & 89.52 & 80.32 & 81.64 & 44.76 & 57.96 & 60.96 & 67.83 \\
OmniGen2 + \textbf{DAR}     & \textbf{89.84} & \textbf{81.00} & \textbf{82.12} & \textbf{48.72} & \textbf{59.28} & 60.73 & \textbf{69.21} \\
\bottomrule
\end{tabular}

\label{tab:performance_comparison}
\end{table*}


\paragraph{Evaluation on Other Benchmarks}

To further verify the effectiveness and generalization of our DAR method, we evaluate it on two multi-image benchmarks, OmniContext~\cite{wu2025omnigen2} and XVerseBench~\cite{chen2025xverse}. As Table~\ref{tab:omnicontext_performance} shows, applying DAR to open-source UMMs (OmniGen2 and BAGEL) consistently improves performance across nearly all single- and multi-subject categories. OmniGen2 gains notably in Character and Object metrics under both Single and Scene settings, while BAGEL improves in most categories.

On XVerseBench (Table~\ref{tab:xversebench_performance}), focused on fine-grained subject identification and attribute similarity, DAR also enhances baseline models. OmniGen2 achieves consistent gains in ID-Sim, IP-Sim, and average scores for Single- and Multi-Subject tasks. BAGEL improves in DPG, ID-Sim, AES, and average metrics. These results demonstrate that DAR robustly enhances multi-image context understanding and fine-grained reasoning across diverse datasets.

\begin{table*}[t]
\centering
\setlength{\tabcolsep}{5.5pt} 
\footnotesize               
\caption{Performance of our method on the OmniContext benchmark.}
\begin{tabular}{lccccccccc}
\toprule
\multirow{2}{*}{\textbf{Method}} & \multicolumn{2}{c}{\textbf{SINGLE}} & \multicolumn{3}{c}{\textbf{MULTIPLE}} & \multicolumn{3}{c}{\textbf{SCENE}} & \multirow{2}{*}{\textbf{Average} $\uparrow$} \\
\cmidrule(lr){2-3} \cmidrule(lr){4-6} \cmidrule(lr){7-9}
& Character & Object       & Character & Object & Char.+Obj. & Character & Object & Char.+Obj. &  \\
\midrule
BAGEL              & 5.713 & 6.223 & 3.028 & 6.904 & 6.802 & 4.244 & 5.157 & 6.219 & 5.536 \\
BAGEL + \textbf{DAR}       & \textbf{6.255} & 6.082 & \textbf{4.135} & \textbf{7.175} & \textbf{6.870} & \textbf{4.775} & 4.838 & \textbf{6.301} & \textbf{5.804}  \\
\midrule
OmniGen2           & 8.184 & 7.328 & 6.564 & 7.994 & 8.039 & 6.871 & 7.898 & 7.378 & 7.532  \\
OmniGen2 + \textbf{DAR}    & \textbf{8.299} & \textbf{8.191} & \textbf{6.644} & \textbf{8.418} & 7.860 & \textbf{7.056} & \textbf{7.965} & \textbf{7.713} & \textbf{7.768}  \\
\bottomrule
\end{tabular}
\label{tab:omnicontext_performance}
\vspace{-0.2cm}
\end{table*}

\paragraph{Evaluation on Other Metrics}
We further assess the effectiveness of our DAR method using multiple complementary metrics that capture different aspects of generation quality and semantic alignment. Table~\ref{tab:metrics_distance_bold} presents comparisons on CLIP~\cite{radford2021learning} scores for both text-to-image (T2I) and image-to-image (I2I) similarity, DINO v2~\cite{oquab2023dinov2} feature similarity (I2I), and perceptual similarity measured by LPIPS~\cite{zhang2018unreasonable} (I2I), evaluated under Object Composition and Spatial Geometric Constraints tasks. For all image-to-image (I2I) similarity measures, we extract object regions described in the prompt from both reference and generated images using GroundingDINO~\cite{liu2024grounding}, ensuring focused and accurate comparison. Both OmniGen2 and BAGEL consistently improve across the majority of metrics with DAR, showing clear gains in semantic alignment and spatial consistency. These improvements demonstrate that DAR enhances multi-object semantics and visual quality effectively, with minimal overhead, confirming its robustness.

\begin{table*}[t]
\centering
\setlength{\tabcolsep}{4pt} 
\footnotesize 
\caption{Performance comparison on the XVerseBench benchmark.}
\begin{tabular}{lcccccccccccc}
\toprule
\multirow{2}{*}{\textbf{Method}} &
\multicolumn{5}{c}{\textbf{Single-Subject}} & 
\multicolumn{5}{c}{\textbf{Multi-Subject}} & 
\multirow{2}{*}{\textbf{Overall} $\uparrow$} \\
\cmidrule(lr){2-6} \cmidrule(lr){7-11}
& DPG & ID-Sim & IP-Sim & AES & AVG (Single) $\uparrow$ 
& DPG & ID-Sim & IP-Sim & AES & AVG (Multi) $\uparrow$ & \\
\midrule
BAGEL             & 93.65 & 3.11 & 46.91 & 47.96 & 47.91 & 89.60 & 2.62 & 32.39 & 45.87 & 42.62 & 45.26 \\
BAGEL + \textbf{DAR}     & \textbf{94.70} & \textbf{3.61} & 47.01 & \textbf{48.83} & \textbf{48.54} 
                   & \textbf{92.13} & \textbf{4.28} & \textbf{33.38} & 45.86 & \textbf{43.91} & \textbf{46.23} \\
\midrule
OmniGen2          & 92.60 & 4.25 & 56.16 & 57.11 & 52.53 & 91.55 & 11.51 & 38.40 & 57.56 & 49.76 & 51.14 \\
OmniGen2 + \textbf{DAR} & \textbf{92.98} & \textbf{5.23} & \textbf{57.66} & 57.08 & \textbf{53.24} 
                   & \textbf{92.56} & \textbf{12.15} & \textbf{39.05} & 57.15 & \textbf{50.23} & \textbf{51.73} \\
\bottomrule
\end{tabular}

\label{tab:xversebench_performance}
\end{table*}

\begin{table*}[t]
\centering
\caption{Evaluation on other metrics, including CLIP score (Text-to-Image and Reference Images-to-Image), DINO v2 (Reference Images-to-Image), and LPIPS (Reference Images-to-Image) under Object Composition and Spatial Geometric Constraints tasks.}
\begin{tabular}{lcccccccc}
\toprule
\multirow{2}{*}{\textbf{Model}} &
\multicolumn{2}{c}{\textbf{CLIP Score (T2I)$\uparrow$}} &
\multicolumn{2}{c}{\textbf{CLIP Score (I2I)$\uparrow$}} &
\multicolumn{2}{c}{\textbf{DINO v2 (I2I)$\uparrow$}} &
\multicolumn{2}{c}{\textbf{LPIPS (I2I)$\downarrow$}} \\
\cmidrule(lr){2-3} \cmidrule(lr){4-5} \cmidrule(lr){6-7} \cmidrule(lr){8-9}
& Object & Spatial & Object & Spatial & Object & Spatial & Object & Spatial \\
\midrule
Nano-Banana      & 0.3505 & 0.3665 & 0.9327 & 0.9382 & 0.8962 & 0.9066 & 0.4934 & 0.4726 \\
GPT-Image      & 0.3572 & 0.3742 & 0.9089 & 0.9168 & 0.8663 & 0.8815 & 0.6431 & 0.6330 \\
Qwen-Image-Edit  & 0.3558 & 0.3700 & 0.9208 & 0.9292 & 0.8789 & 0.8923 & 0.5542 & 0.5249 \\
UNO              & 0.3391 & 0.3566 & 0.9007 & 0.9073 & 0.8301 & 0.8366 & 0.6453 & 0.6486 \\
DreamOmni2       & 0.3454 & 0.3647 & 0.9133 & 0.9142 & 0.8719 & 0.8787 & 0.6217 & 0.6244 \\
\midrule
BAGEL            & 0.3586 & 0.3748 & 0.9155 & 0.9147 & 0.8766 & 0.8783 & 0.6073 & 0.6210 \\
BAGEL + \textbf{DAR}       & \textbf{0.3612} & \textbf{0.3761} & \textbf{0.9201} & \textbf{0.9172} & \textbf{0.8828} & \textbf{0.8821} & \textbf{0.6018} & \textbf{0.6207} \\
OmniGen2         & 0.3646 & 0.3828 & 0.9102 & 0.9139 & 0.8742 & 0.8751 & 0.6373 & 0.6389 \\
OmniGen2 + \textbf{DAR}    & \textbf{0.3648} & 0.3828 & \textbf{0.9130} & \textbf{0.9158} & \textbf{0.8757} & \textbf{0.8778} & \textbf{0.6327} & \textbf{0.6360} \\
\bottomrule
\end{tabular}

\label{tab:metrics_distance_bold}
\end{table*}

\subsection{Ablation Analysis}

\begin{table}[t]
\centering
\small 
\caption{Analysis of the impact of the numbers of reference images on UMMs in the MICON-Bench Object Composition Task.}
\label{tab:ref_image_ablation}
\begin{tabular}{lcccc}
\toprule
\textbf{Model} & \textbf{Ref=2} & \textbf{Ref=3} & \textbf{Ref=4} & \textbf{Ref=5} \\
\midrule
BAGEL          & 88.50          & 84.37          & 75.11          & 66.36          \\
OmniGen2       & 92.18          & 89.52          & 74.92          & 67.00          \\
\bottomrule
\end{tabular}
\vspace{-0.2cm}
\end{table}

\paragraph{Impact of Different Numbers of Reference Images on UMMs' Generation Performance.}

We perform an ablation study to analyze how varying the number of reference images affects the performance of the UMMs on the MICON-Bench Object Composition Task. In particular, we extended the original task settings by adding configurations with 4 and 5 reference images, forming a dedicated subset for this analysis. As summarized in Table~\ref{tab:ref_image_ablation}, both BAGEL and OmniGen2 were evaluated using reference image counts ranging from 2 to 5.
The results clearly show a decreasing trend in performance as the number of reference images increases. This suggests that fusing information from more references imposes greater challenges on UMMs, leading to a decline in generation quality. The observed performance degradation points to the need for improved fusion techniques and robustness strategies in future models when handling complex multi-reference scenarios.

\begin{figure}[t]
  \centering
  \includegraphics[width=0.98\linewidth]{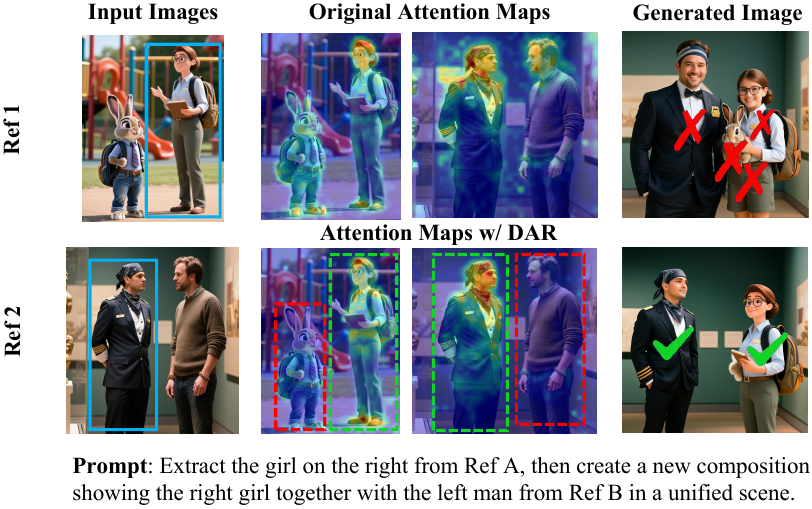}
  \caption{
DAR suppresses noisy, irrelevant attention (\textcolor{red}{red boxes}) and re-focuses activations on the correct subjects (\textcolor{green}{green boxes}), resulting in cleaner referencing and more faithful composition. 
The \textcolor{blue}{blue boxes} highlight the target subjects. 
}
  \label{fig:vis2}
\vspace{-0.2cm}
\end{figure}

\paragraph{Qualitative Effect of Attention Rebalancing.}
As shown in Figure~\ref{fig:vis2}, the original attention maps tend to scatter across unrelated background areas or unintended persons, causing identity confusion and hallucinated details in the generated image. By rebalancing attention, DAR suppresses these spurious activations and better aligns attention with the intended reference subjects, resulting in improved localization and more faithful composition.

\subsection{Visualization}
Figure~\ref{fig:vis} shows the qualitative advantages of our method over baseline models across all tasks in MICON-Bench. While existing models often fail to integrate information from multiple references—leading to missing objects, incorrect spatial arrangement, attribute mixing, inconsistent component transfer, foreground–background artifacts, or incoherent temporal reasoning—our Dynamic Attention Rebalancing (DAR) consistently improves cross-image coherence and fine-grained visual alignment. By suppressing irrelevant attention and reinforcing reference-relevant regions, our method enables accurate object inclusion, precise spatial placement, clean attribute disentanglement, reliable component transfer, seamless foreground replacement, and more plausible story continuation. These visual results corroborate our quantitative findings, demonstrating that DAR enhances multi-image understanding and significantly mitigates cross-image hallucination issues.

\section{Conclusion}
In this work, we introduce MICON-Bench, a comprehensive benchmark targeting multi-image context generation—a challenging yet essential capability for next-generation multimodal generative models. Our benchmark covers six diverse task categories and provides an automatic Evaluation-by-Checkpoint framework that systematically measures semantic and visual consistency. To further advance this emerging direction, we propose Dynamic Attention Rebalancing (DAR), a training-free and plug-and-play mechanism that adaptively redistributes attention across reference images during inference. Extensive experiments demonstrate that DAR effectively enhances cross-image coherence and reduces hallucinations across a wide range of UMMs. We believe MICON-Bench and DAR together provide a foundation for multi-image reasoning and the development of reliable generative systems.

\section*{Acknowledgments}
This work was supported by the National Science Fund for Distinguished Young Scholars (No.62025603), the National Natural Science Foundation of China (No. U22B2051, No. 62302411) and China Postdoctoral Science Foundation (No. 2023M732948), and the Zhongguancun Academy, Beijing, China (No. 20240103).
{
    \small
    \bibliographystyle{ieeenat_fullname}
    \bibliography{main}
}

\clearpage
\setcounter{page}{1}
\maketitlesupplementary

\section{Limitations}
Despite the promising results, our current evaluation pipeline inherently relies on the underlying MLLM’s perception capability, which is itself susceptible to hallucinations. Consequently, erroneous model predictions may propagate into the evaluation process, potentially biasing the reported performance. Moreover, the effectiveness of the proposed DAR-based re-weighting mechanism remains constrained by the quality of the original attention maps. When the base model fails to correctly interpret the reference image—e.g., missing fine-grained semantics or mis-localizing key regions—the re-weighted attention may still amplify incorrect signals rather than rectify them.

\section{Benchmark Detail}
\label{sec:rationale}
\subsection{Data Statistics}
\begin{figure*}[htbp]
  \centering
  \includegraphics[width=0.86\textwidth]{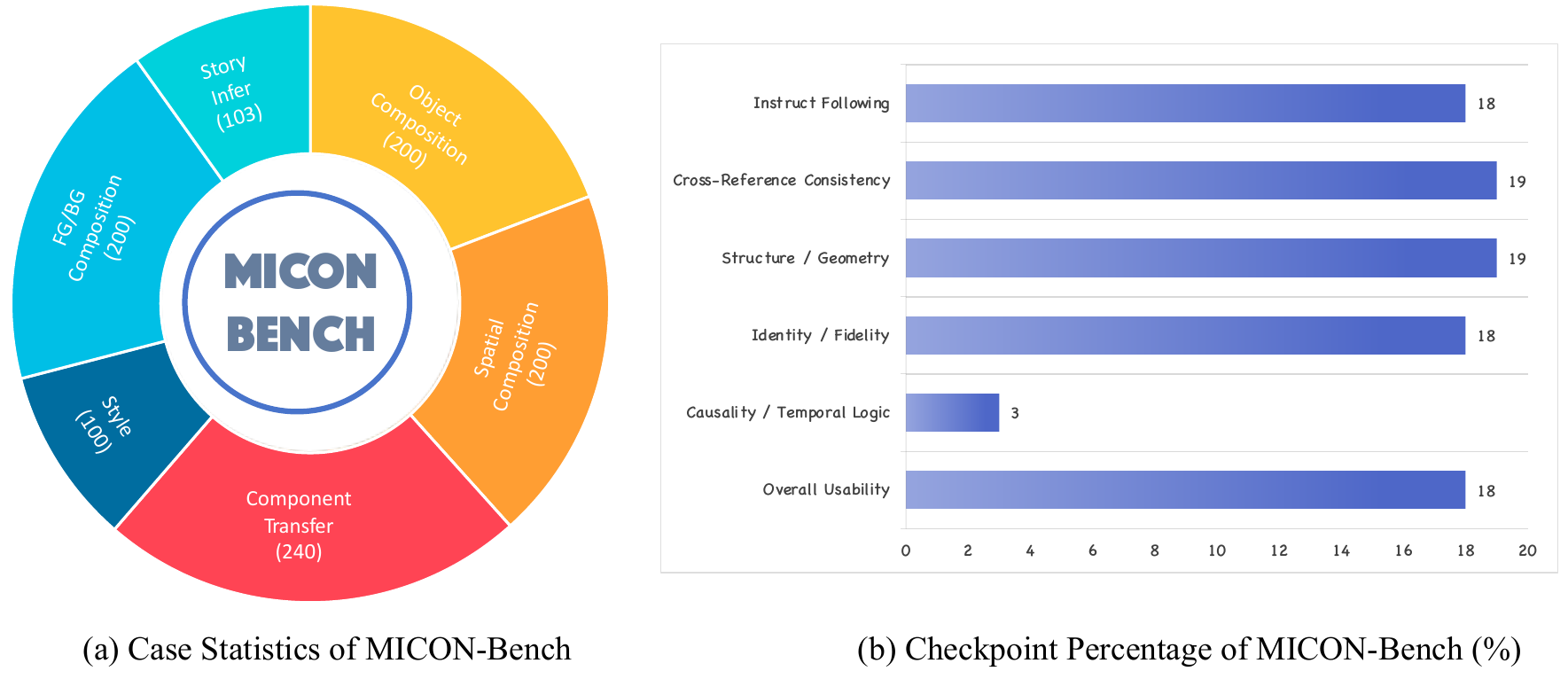}
  \caption{
The data statistics of MICON-Bench.
}
  \label{fig:sta}
\end{figure*}

\paragraph{Dataset Composition.}
MICON-Bench contains six types of multi-image context generation tasks: Object Composition, Spatial Composition, Attribute Disentanglement, Component Transfer, Foreground/Background (FG/BG) Composition, and Story Inference.
The per-task number of cases is visualized in Figure~\ref{fig:sta}~(a).
Beyond case counts, we further summarize the total number of images and how many cases use two or three reference images, as shown in Table~\ref{tab:data_statistics}.
Overall, the benchmark comprises 1,043 cases and 2,518 images, among which 611 cases use two reference images and 432 cases use three reference images.

\begin{table}[htbp]
\small
\centering
\caption{Statistics of MICON-Bench, including the number of cases, images, and the distribution of two- and three-reference-image settings for each task.}
\label{tab:data_statistics}
\begin{tabular}{lcccc}
\toprule
\textbf{Task} & \textbf{\#Cases} & \textbf{\#Images} & \textbf{\#2 Ref} & \textbf{\#3 Ref} \\
\midrule
Object   & 200 & 482 & 118 & 82  \\
Spatial  & 200 & 498 & 102 & 98  \\
Attribut & 100 & 300 & 0   & 100 \\
Component & 240 & 601 & 119 & 121 \\
FG/BG    & 200 & 400 & 200 & 0   \\
Story    & 103 & 237 & 72  & 31  \\
\midrule
\textbf{Total} & \textbf{1,043} & \textbf{2518} & \textbf{611} & \textbf{432} \\
\bottomrule
\end{tabular}
\end{table}

\paragraph{Evaluation Checkpoints.}
For automatic evaluation, we adopt the Evaluation-by-Checkpoint paradigm described in the main paper, where an MLLM verifies whether each generated image satisfies a set of fine-grained visual and semantic conditions.
For each task, we design five evaluation dimensions (e.g., instruction following, identity preservation, structural consistency, cross-reference consistency, and overall usability) and instantiate several binary checkpoints under each dimension.
Figure~\ref{fig:sta}~(b) summarizes the percentage of checkpoints per dimension for all six tasks.


\subsection{Benchmark Construction}
\paragraph{Image Generation.}  
We employ Qwen-Image to generate image data for five fundamental compositional tasks: Object Composition, Spatial Composition, Attribute Disentanglement, Component Transfer, and Foreground/Background (FG/BG) Composition.
We first construct comprehensive collections of subjects, attributes, spatial relations, and scenes.
From these collections, we sample elements to fill the reference image prompt templates, which guide Qwen-Image in generating the reference images for each task. Simultaneously, we use the same sampled elements to populate the task prompt templates, thereby forming the task prompts used in our MICON-Bench.
This approach ensures that the reference images and corresponding task prompts are naturally aligned and rooted in a shared, well-defined compositional space, providing consistency between the data generation and evaluation processes.
The prompt templates for each task are shown in Table~\ref{tab:prompt_templates}.

For the \textbf{Story Inference} task, we construct a dataset of short visual narratives that require causal reasoning over multiple reference images. The visual stories are categorized into three types: (1) \textit{physical property changes} (e.g., a candle being blown out), (2) \textit{causal commonsense} (e.g., everyday cause--effect scenarios), and (3) \textit{action continuation} (e.g., predicting the next step of an ongoing action).

For each accepted script, we generate a four-panel comic using image generation models and manually crop each panel to obtain clean reference images. We further apply DiffBIR-based restoration to enhance sharpness and reduce artifacts, resulting in the final Story Inference reference sets used in our benchmark. The model is then asked to infer a plausible continuation of the story given two or three reference images.

\paragraph{Data Filtering.}
To ensure high-quality and semantically consistent dataset construction, we employ a two-stage filtering pipeline that combines automatic and manual reviews. First, a Multimodal Large Language Model (MLLM) is utilized to perform coarse-grained automatic filtering by verifying whether generated images faithfully adhere to the corresponding textual prompts. The MLLM evaluates attributes such as object completeness, spatial relationships, style consistency, and component correctness, effectively identifying blatant failures or inconsistencies at scale. Subsequently, filtered samples undergo meticulous human inspection to catch subtle issues beyond automated detection, including unnatural compositions, visual artifacts, and nuanced spatial or attribute violations. This dual-layer filtering protocol helps ensure that the final dataset achieves both diversity and high fidelity, providing a reliable benchmark for compositional visual understanding and generation.

\subsection{Evaluation Detail}
\paragraph{Checkpoints Generation.} 
We adopt a fine-grained, checkpoint-based evaluation protocol powered by an MLLM judge. Across all tasks, we define seven evaluation dimensions:
\begin{itemize}
    \item \textbf{A. Instruction Following} -- whether the generated image follows the textual instruction and task specification.
    \item \textbf{B. Identity / Fidelity} -- whether object identities and key attributes match their references.
    \item \textbf{C. Structure / Geometry} -- whether spatial layout, geometry, and physical plausibility are preserved.
    \item \textbf{D. Cross-Reference Consistency} -- whether information from multiple reference images is integrated without contradiction.
    \item \textbf{E. Causality} -- whether temporal progression and cause--effect relations are reasonable.
    \item \textbf{F. Text Grounding} -- whether textual content (e.g., overlaid text) is correctly rendered and grounded.
    \item \textbf{G. Overall Usability} -- whether the final image is natural, coherent, and visually usable.
\end{itemize}

Different tasks activate different subsets of these dimensions according to their nature (e.g., Foreground/Background Composition focuses on A/B/C/D/G, while Story Inference emphasizes A/B/C/D/E/G). Within each active dimension, we design $2$--$4$ concrete \emph{checkpoints} that specify what the MLLM should verify. For each test case, we feed the instructions, reference images, and the generated image into the MLLM, and ask it to decide, for every checkpoint, whether the requirement is satisfied (pass/fail) along with a short justification.

We further mark one key checkpoint per dimension as a \textbf{hard constraint}, denoted by $(\mathrm{H}x)$ (e.g., A\_check\_1 (H1), B\_check\_1 (H2)). If a hard-constraint checkpoint is judged as failed, the score of that dimension is capped at $0.4$ (on a $[0,1]$ scale), regardless of other checkpoints in that dimension. For each sample, we convert the proportion of satisfied checkpoints in each dimension into a dimension score in $[0,1]$, aggregate over all active dimensions, and linearly rescale the result to obtain a final score in $[0,100]$.

\vspace{0.5em}
\paragraph{Example of Checkpoints.}
The checkpoints of the Object Composition task are shown in Table~\ref{tab:eval_guidelines}. Other tasks (e.g., Spatial Geometric Constraints, Foreground/Background Composition, Story Inference, and Text-based Editing) are defined analogously, each with a tailored subset of dimensions and a small set of task-specific checkpoints under the same unified framework.

\paragraph{Evaluation of Story Generation.}
For the complex Story Inference task, which involves causal reasoning over multiple images, we adopt a hybrid evaluation scheme that combines traditional MLLM checkpoint scoring with a human-constructed answer set. Concretely, the final score is a weighted sum of two components: (i) the standard checkpoint-based MLLM score described above (40\%), and (ii) a human-answer-set score (60\%).

\begin{figure}[t]
  \centering
  \includegraphics[width=0.98\linewidth]{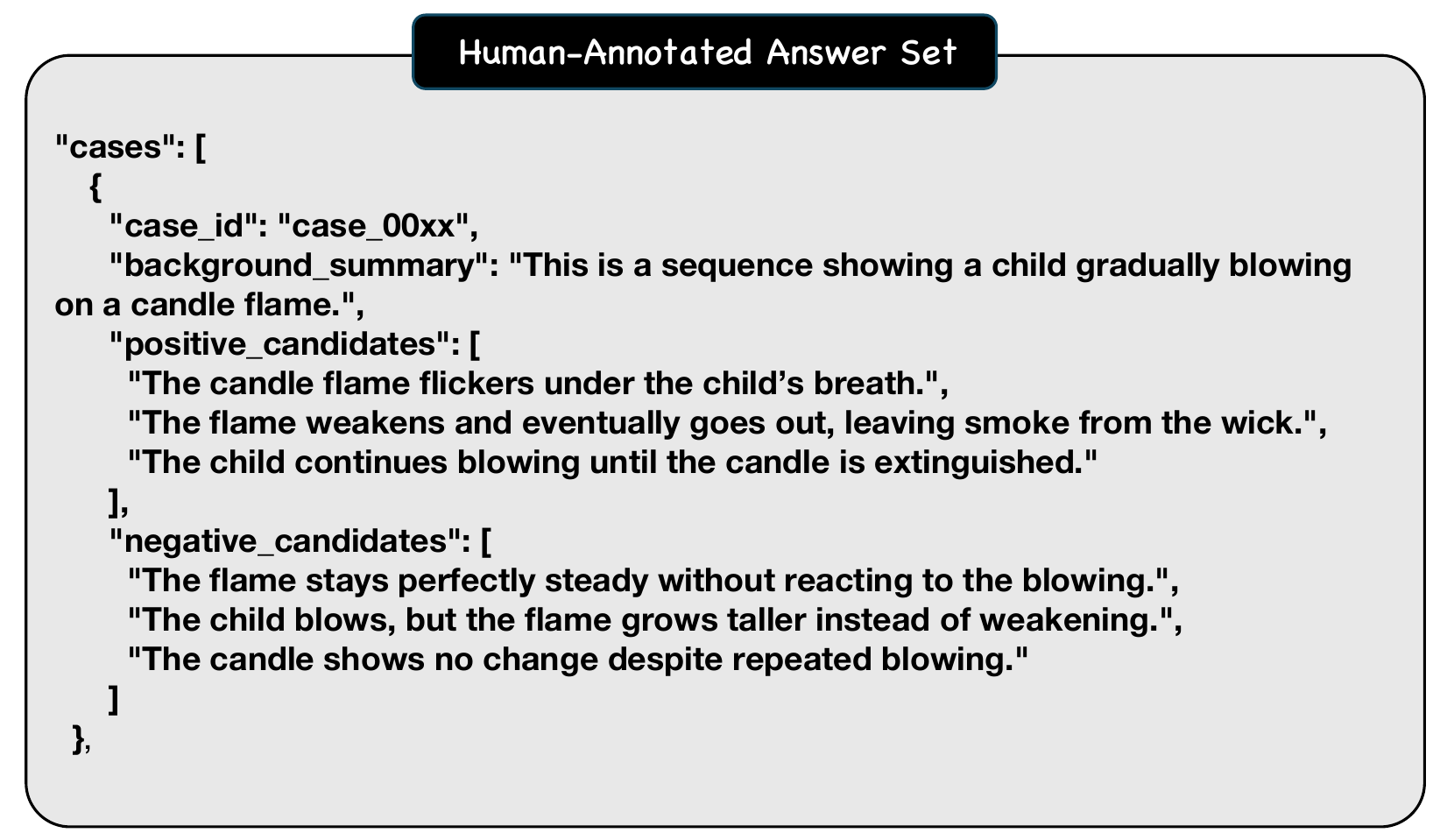}
  \caption{
Human-annotated answer set for the Story Inference task. Each example pairs the reference image sequence with canonical human-written descriptions of what should happen next, which are used as targets for MLLM-based evaluation.
}
  \label{fig:sup_story}
\end{figure}

For the human-answer-set component, we construct a small set of human-written targets for each story, including a concise narrative description, plausible predictions of what is likely to happen next, and counterfactual outcomes that are unlikely to occur. During evaluation, we feed the MLLM with the reference images, the model-generated continuation image, and the corresponding human answer set, and ask it to assess how well the generated image matches these canonical outcomes in terms of causal progression, visual evidence, and overall story resolution. An illustration of the human answer set and its usage in the evaluation prompt is shown in Fig.~\ref{fig:sup_story}.

This hybrid design leverages human-authored answers as semantically precise, high-level anchors for what should happen next, reducing ambiguity and overly lenient judgments compared to using only low-level checkpoints. At the same time, combining checkpoint-based verification with answer-set matching encourages models to be accurate both at the local level and at the global level of causal consistency and narrative coherence, yielding a more robust and faithful evaluation of story generation quality.

\section{Additional Experiments}
\subsection{Ablation Study}
\paragraph{Inference Cost.}
Table~\ref{tab:inference_time_compact} presents the average inference time measured on the Object Composition dataset using a single NVIDIA A800-40GB GPU.  
Our method (+OURS) introduces only minimal overhead compared to the respective baselines for both BAGEL and Omnigen2 across different numbers of reference images.  
Specifically, the runtime increase remains within a small margin (approximately 5–10\%), demonstrating that the performance gains come at a low computational cost.  
This demonstrates the efficiency of our approach and its practicality for real-world deployment, where inference speed is critical.  

\begin{table}[htbp]
\centering
\small
\caption{Average inference time (seconds) on Object Composition dataset measured on a single NVIDIA A800-40GB GPU.}
\label{tab:inference_time_compact}
\setlength{\tabcolsep}{5pt} 
\begin{tabular}{@{}lcccc@{}}
\toprule
\# Reference Images & \multicolumn{2}{c}{BAGEL} & \multicolumn{2}{c}{Omnigen2} \\ \cmidrule(lr){2-3} \cmidrule(lr){4-5}
                   & Baseline     & +OURS      & Baseline     & +OURS     \\ \midrule
2                  & 57.13        & 61.70      & 72.01        & 74.71     \\
3                  & 69.72        & 77.01      & 97.80        & 101.58    \\ \bottomrule
\end{tabular}
\end{table}

\paragraph{Ablation of Weight Factor $\gamma$.}
Table~\ref{tab:gamma_ablation} presents the performance of our model across six subtasks of MICON-Bench under varying weight factors \(\gamma\).  
The results indicate that \(\gamma = 0.15\) achieves the best overall performance, obtaining the highest accuracy in Object, Spatial, FG/BG, and Story tasks.  
While a smaller \(\gamma = 0.05\) yields competitive results in Attribute and Spatial tasks, it underperforms notably in FG/BG and Story.  
As \(\gamma\) increases beyond 0.15, the performance across all tasks drops sharply, suggesting that overly strong weighting negatively impacts learning.  
Notably, at \(\gamma=0.55\), all task accuracies fall to very low levels, indicating that the model struggles to generalize due to excessive regularization.  
These findings demonstrate the importance of carefully tuning \(\gamma\) to balance the contributions of different components during training, ensuring robust performance across diverse subtasks.
\begin{table}[htbp]
\centering
\small
\caption{Ablation study on the effect of different weight factor $\gamma$.}
\label{tab:gamma_ablation}
\setlength{\tabcolsep}{4pt}
\begin{tabular}{c|cccccc}
\toprule
$\gamma$ & Object & Spatial & Attribute & Component & FG/BG & Story \\
\midrule
0.05       & 86.11 & 91.68 & \textbf{92.24} & \textbf{56.07} & 59.29 & 64.75 \\
0.15 & \textbf{88.04} & \textbf{91.88} & 90.76 & 56.06 & \textbf{71.24} & \textbf{66.34} \\
0.25       & 76.58 & 85.41 & 75.43 & 44.36 & 64.41 & 62.86 \\
0.35       & 53.36 & 45.78 & 37.84 & 23.76 & 65.14 & 58.96 \\
0.55       & 8.39  & 18.30 & 0.67  & 8.39  & 22.77 & 40.44 \\
\bottomrule
\end{tabular}
\end{table}

\begin{table*}[t]
\centering
\caption{Ablation study on prompt sensitivity check for the Evaluation-by-Checkpoint.}
\begin{tabular}{lccccccc}
\toprule
\textbf{Model} & \textbf{Object} & \textbf{Spatial} & \textbf{Attribute} & \textbf{Component} & \textbf{FG/BG} & \textbf{Story} & \textbf{Avg. Score} \\
\midrule

BAGEL(Generic)        & 87.72 & 84.49 & 87.51 & 47.33 & 51.78 & 63.66 & 68.52 \\
BAGEL(Specific)           & 88.17 & 84.09 & 85.20 & 46.09 & 51.78 & 63.51 & 68.01 \\

\midrule

BAGEL+\textbf{DAR}(Generic) & 91.22 & 87.81 & 89.24 & 48.49 & 78.22 & 68.86 & 75.84 \\
BAGEL+\textbf{DAR}(Specific) & 91.72 & 87.53 & 89.42 & 47.67 & 80.00 & 67.51 & 75.92 \\

\bottomrule
\end{tabular}

\label{tab:prompt sensitivity check}
\end{table*}

\begin{table*}[t]
\centering
\caption{Performance comparison of different models on our MICON-Bench evaluated by InternVL3.5-38B.}
\begin{tabular}{lccccccc}
\toprule
\textbf{Model} & \textbf{Object} & \textbf{Spatial} & \textbf{Attribute} & \textbf{Component} & \textbf{FG/BG} & \textbf{Story} & \textbf{Avg. Score} \\
\midrule

Nano-Banana~\cite{google2025gemini}         & 97.88 & 98.18 & 72.03 & 79.41 & 83.71 & 79.06 & 86.63 \\
GPT-Image~\cite{hurst2024gpt}               & 98.85 & 97.25 & 83.70 & 85.26 & 91.08 & 81.58 & 90.77 \\
UNO~\cite{wu2025less}                       & 62.30 & 57.75 & 55.11 & 27.56 & 28.47 & 38.17 & 43.87 \\
DreamOmni2~\cite{xia2025dreamomni2}         & 88.95 & 81.69 & 62.64 & 65.52 & 81.01 & 59.94 & 75.26 \\
Qwen-Image-Edit-2507~\cite{wu2025qwen}      & 98.03 & 97.57 & 65.05 & 57.12 & 74.95 & 60.74 & 77.26 \\
\midrule
BAGEL~\cite{deng2025emerging}               & 89.08 & 89.32 & 69.05 & 59.57 & 72.39 & 54.32 & 73.78 \\
BAGEL + \textbf{DAR}                        & \textbf{89.56} & \textbf{91.57} & \textbf{72.39} & \textbf{66.05} & \textbf{80.33} & \textbf{60.93} & \textbf{78.29} \\
\midrule
OmniGen2~\cite{wu2025omnigen2}              & 88.77 & 81.91 & 69.13 & 47.53 & 69.98 & 57.37 & 69.38 \\
OmniGen2 + \textbf{DAR}                     & \textbf{89.50} & \textbf{83.06} & \textbf{70.71} & \textbf{49.79} & \textbf{72.26} & \textbf{61.15} & \textbf{71.22} \\
\bottomrule
\end{tabular}

\label{tab:internvl_full}
\end{table*}

\begin{table*}[t]
\centering
\caption{\textbf{Human Evaluation} results on the uniformly sampled subset of the MICON-Bench.}
\begin{tabular}{lccccccc}
\toprule
\textbf{Model} & \textbf{Object} & \textbf{Spatial} & \textbf{Attribute} & \textbf{Component} & \textbf{FG/BG} & \textbf{Story} & \textbf{Avg. Score} \\
\midrule

Nano-Banana-Pro~\cite{google2025gemini}                              & 90.00 & 93.20 & 92.50 & 85.31 & 95.29 & 87.33 & 90.61 \\
Nano-Banana~\cite{google2025gemini}         & 92.45 & 90.34 & 89.23 & 76.35 & 87.67 & 75.00 & 85.17 \\
GPT-Image~\cite{hurst2024gpt}               & 92.90 & 93.46 & 91.73 & 79.80 & 87.12 & 90.00 & 89.17 \\
UNO~\cite{wu2025less}                       & 56.03 & 54.12 & 60.10 & 15.73 & 25.00 & 40.09 & 41.84 \\
DreamOmni2~\cite{xia2025dreamomni2}         & 90.00 & 81.25 & 83.30 & 51.98 & 78.20 & 60.35 & 74.18 \\
Qwen-Image-Edit-2507~\cite{wu2025qwen}      & 89.40 & 84.30 & 75.00 & 44.76 & 77.95 & 59.50 & 71.82 \\
\midrule
BAGEL~\cite{deng2025emerging}               & 88.24 & 84.54 & 63.56 & 54.12 & 70.67 & 65.30 & 71.07 \\
BAGEL + \textbf{DAR}                        & \textbf{89.67} & \textbf{88.36} & \textbf{63.45} & \textbf{58.34} & \textbf{74.95} & \textbf{69.73} & \textbf{74.08} \\
\midrule
OmniGen2~\cite{wu2025omnigen2}              & 86.20 & 73.89 & 52.45 & 41.03 & 60.17 & 56.79 & 61.75 \\
OmniGen2 + \textbf{DAR}                     & \textbf{88.76} & \textbf{75.57} & \textbf{54.35} & \textbf{42.00} & \textbf{64.07} & \textbf{58.56} & \textbf{63.88} \\
\bottomrule
\end{tabular}

\label{tab:human_detail}
\end{table*}

\begin{table}[t]
    \centering
    \caption{\textbf{Reliability Analysis} }
    \resizebox{\linewidth}{!}{
    \begin{tabular}{l|c|cc}
        \toprule
        \multirow{2}{*}{\textbf{Target Model}} & \textbf{Human} & \multicolumn{2}{c}{\textbf{Automatic Verifiers}}  \\
         & \textbf{Score} & \textbf{Qwen3-VL} & \textbf{InternVL3.5}  \\
        \midrule
        Nano-Banana-pro & 90.61 & 92.42 & 90.93  \\
        GPT-Image & 89.17 & 89.15 & 89.05  \\
        Nano-Banana & 85.17 & 85.68 & 85.72  \\
        \textbf{BAGEL + DAR (Ours)} & \textbf{74.08} & \textbf{75.62} & \textbf{75.57}  \\
        DreamOmni2 & 74.18 & 74.58 & 73.36  \\
        BAGEL & 71.07 & 71.42 & 71.08 \\
        Qwen-Image-Edit & 71.82 & 72.26 & 72.35  \\
        \textbf{OmniGen2 + DAR (Ours)} & \textbf{63.88} & \textbf{64.58} & \textbf{63.32}  \\
        OmniGen2 & 61.75 & 62.44 & 61.12  \\
        UNO & 41.84 & 41.62 & 41.44  \\
        \midrule
        \textit{Avg. Deviation from Human} & - & \textit{+0.67} & \textit{+0.54}   \\
        \bottomrule
    \end{tabular}
    }
    \vspace{-0.4cm}
    \label{tab:human_alignment_summary}
\end{table}

\subsection{Prompt Sensitivity Check}
To rigorously validate the reliability of our Evaluation-by-Checkpoint framework, we investigate whether the MLLM-based verifier is overly sensitive to the specific phrasing of the evaluation checkpoints.
We conduct a prompt sensitivity ablation study using the BAGEL model. We randomly sample a subset of 180 cases from the MICON-Bench dataset, ensuring an even distribution across all six task categories (30 cases per task). For each case, we design two distinct sets of evaluation checkpoints:
\begin{itemize}
    \item \textbf{Generic Checkpoints:} These utilize abstract, template-based descriptions (e.g., \textit{``Does the image include all specified objects?''}).
    \item \textbf{Specific Checkpoints:} These replace abstract terms with concrete, instance-level entities derived directly from the user prompt (e.g., \textit{``Does the image include a giraffe and a wooden chair?''}).
\end{itemize}
The quantitative results are summarized in Table ~\ref{tab:prompt sensitivity check}. The performance of BAGEL evaluated under Generic checkpoints is 68.52, while the score under Specific checkpoints is 68.01. This high degree of consistency across all six tasks demonstrates that our chosen MLLM verifier exhibits robust reasoning capabilities. It successfully comprehends the core intent of the evaluation dimensions without being biased by phrasing variations, thereby confirming that our automatic evaluation metric is both stable and reliable.

\subsection{Verifier Generalization}
To demonstrate the robust generalization of our Evaluation-by-Checkpoint framework, we replace our default verifier, Qwen3-VL-32B-Instruct, with another state-of-the-art open-source MLLM, InternVL3.5-38B. We re-evaluate all generated images across the entire MICON-Bench dataset (all cases, without sampling) using the exact same checkpoints.

The results are presented in Table \ref{tab:internvl_full}. As the results show, despite changing the verifier, the relative performance rankings of all evaluated models remain perfectly consistent with our main results. This comprehensive evaluation on the full dataset proves that our framework successfully captures fundamental semantic and visual alignments, rather than idiosyncratic biases of a single MLLM, demonstrating strong robustness to the choice of verifier.

\subsection{Human Alignment Check}
To ensure that our automated Evaluation-by-Checkpoint metric serves as a reliable proxy for actual human perception, we conduct a rigorous human alignment study. We randomly sample a subset of 120 cases from MICON-Bench, ensuring a strictly balanced distribution of 20 cases for each of the six tasks. Three expert human annotators are instructed to manually evaluate the generated images. To ensure a fair comparison, the human annotators are provided with the exact same binary, hard-constraint checkpoints (e.g., identity preservation, spatial relationships) that were fed to the MLLM verifier. 

The human evaluation scores on the sampled subset are reported in Table \ref{tab:human_detail}. And the Table \ref{tab:human_alignment_summary} provides a comprehensive comparison between human judgments and our default verifier. As illustrated in the summary table, the MLLM evaluator achieves exceptionally high consistency with human judgments. The average deviation from human scores across the models is marginal (only $+0.67$). The relative rankings of the models provided by the automated verifier perfectly mirror the human rankings. These findings confirm that our hard-constraint rubric effectively bridges the gap between automatic metrics and human visual perception, validating the reliability of our benchmark.

\subsection{Stability and Reproducibility of the MLLM Evaluator}
To ensure that our Evaluation-by-Checkpoint framework serves as a robust scientific metric, we  conduct a stability test by repeatedly evaluating the generated images of a representative model (BAGEL) using our default verifier, Qwen3-VL. Specifically, we performed five independent evaluation runs on the images generated by the BAGEL model. The final average scores for BAGEL across the five trials are remarkably consistent: Run 1 yielded 68.47; Run 2: 68.51; Run 3: 68.48; Run 4: 68.51; and Run 5: 68.47.
The maximum score discrepancy across all five trials is a mere 0.04. This exceptionally low variance indicates that our strictly defined, binary-choice checkpoint mechanism effectively constrains the generative randomness typically associated with MLLMs.

\subsection{More Examples}
\begin{figure*}[t]
  \centering
  \includegraphics[width=0.86\textwidth]{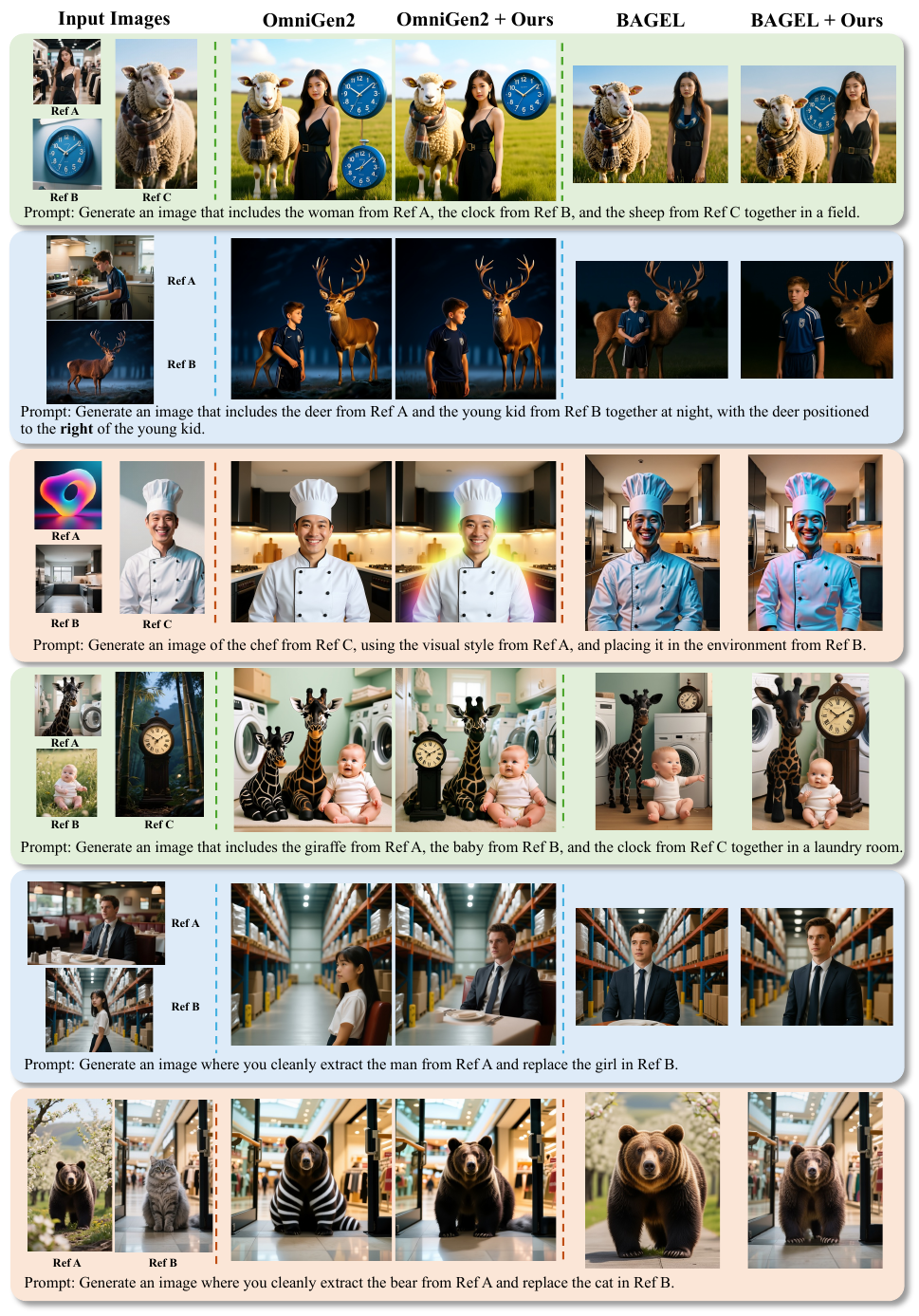}
  \caption{
More visualization examples of our method vs. baseline on MICON-Bench.
}
  \label{fig:sup2}
\end{figure*}

\begin{figure*}[t]
  \centering
  \includegraphics[width=0.86\textwidth]{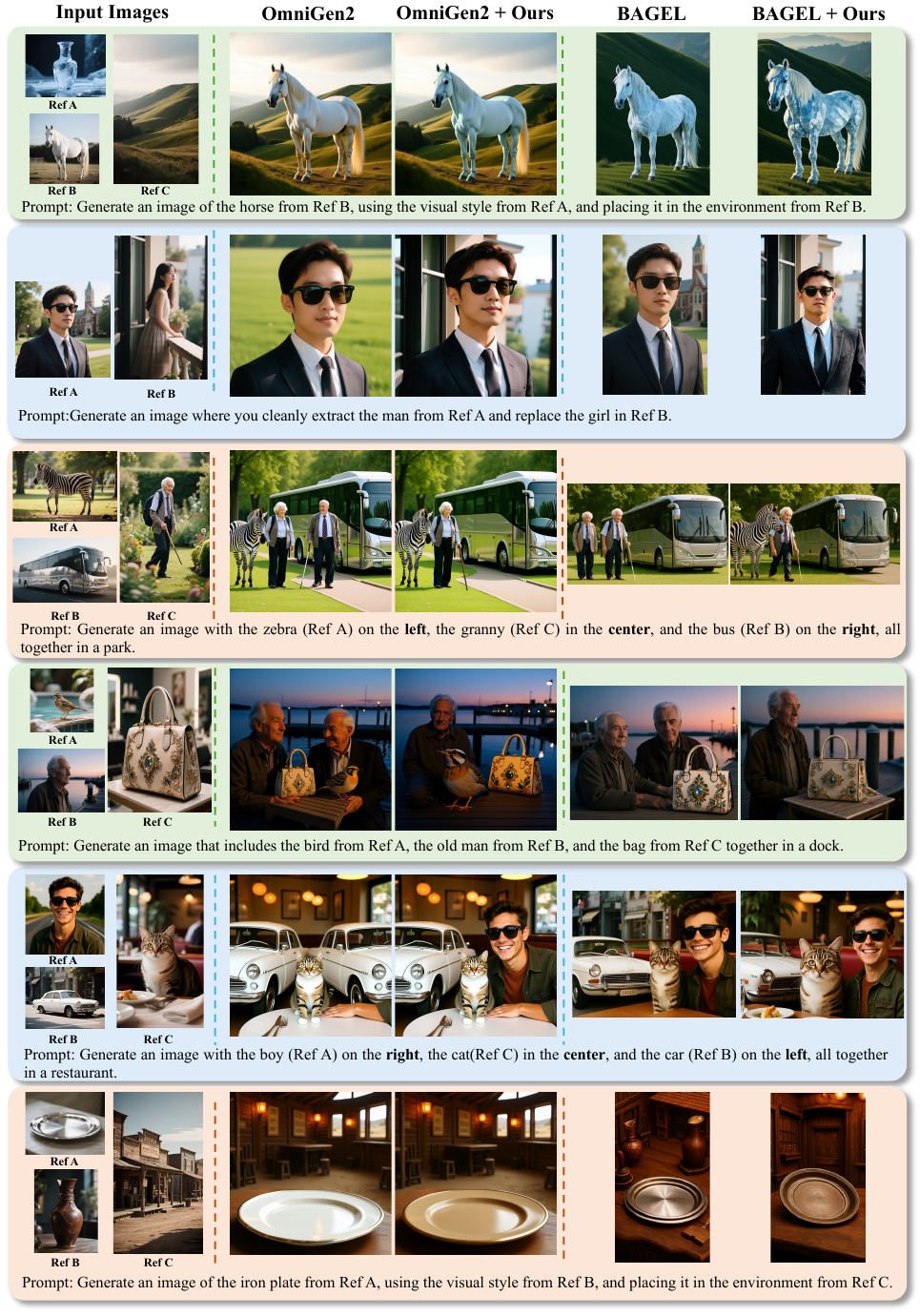}
  \caption{
More visualization examples of our method vs. baseline on MICON-Bench.
}
  \label{fig:sup3}
\end{figure*}

We provide two additional visual examples that further demonstrate the robustness and generalizability of our Dynamic Attention Rebalancing (DAR) approach across diverse multi-image composition scenarios. As shown in Figure~\ref{fig:sup2} and Figure~\ref{fig:sup3}, these examples reinforce DAR’s capability to precisely integrate multiple reference inputs, effectively suppressing irrelevant attention while emphasizing critical details from each source image. As shown, our method successfully avoids common pitfalls seen in baseline models, such as attribute leakage and misplaced spatial arrangements, by maintaining clear object boundaries and consistent attribute rendering. The supplemental figures also highlight DAR’s flexibility in handling various composition challenges, including complex foreground-background interactions and nuanced component transfers. Collectively, these extended visualizations provide compelling qualitative evidence supporting the scalability and effectiveness of DAR in resolving cross-image inconsistencies and achieving fine-grained visual alignment in multi-image tasks.

\begin{table*}[t]
    \centering
    \small
    \caption{\textbf{Prompt Templates for Benchmark Construction.} We detail the reference image prompts and the specific task prompts used for generation, covering object composition, spatial arrangement, attribute disentanglement, component transfer, background replacement, and story inference.}
    \label{tab:prompt_templates}
    \renewcommand{\arraystretch}{1.2}
    \begin{tabular}{l p{3.5cm} p{10cm}}
        \toprule
        \textbf{Task Category} & \textbf{Configuration} & \textbf{Prompt Template} \\
        \midrule
        
        \multirow{3}{*}{\textbf{Object Composition}} 
        & Reference Image & ``A photo of \{personalized\_obj\} in \{scene\}." \\
        \cmidrule{2-3} 
        & Task (2 Refs) & ``Generate an image that contains both the complete \{obj\_a\} and \{obj\_b\} together in \{chosen\_scene\}." \\
        & Task (3 Refs) & ``Generate an image that contains all three objects (\{obj\_a\}, \{obj\_b\}, and \{obj\_c\}) together in \{chosen\_scene\}." \\
        \midrule
        
        \multirow{3}{*}{\textbf{Spatial Composition}} 
        & Reference Image & ``A photo of \{personalized\_obj\} in \{scene\}." \\
        \cmidrule{2-3} 
        & Task (2 Refs) & ``Generate an image that contains both the complete \{obj\_a\} and \{obj\_b\} together in \{chosen\_scene\}, with \{obj\_a\} positioned to the \{spatial\_relation\} of \{obj\_b\}." \\
        & Task (3 Refs) & ``Generate an image that contains all three objects (\{obj\_a\}, \{obj\_b\}, and \{obj\_c\}) together in \{chosen\_scene\}, with \{left\_obj\} on the left, \{center\_obj\} in the center, and \{right\_obj\} on the right." \\
        \midrule
        
        \multirow{4}{*}{\parbox{3cm}{\textbf{Attribute}\\ \textbf{Disentanglement}}} 
        & Ref A (Subject) & ``A photo of a clear \{personalized\_description\} in \{background\_desc\}." \\
        & Ref B (Style) & ``A photo of a \{style\_object\} rendered in \{style\_desc\}." \\
        & Ref C (Background) & ``A photo of beautiful \{specific\_background\}, empty scene without main objects." \\
        \cmidrule{2-3} 
        & Task (A+B+C) & ``Generate an image of the \{main\_object\} from image A, using the visual style from image B, and placing it in the \{specific\_background\} environment from image C." \\
        \midrule
        
        \multirow{4}{*}{\parbox{3cm}{\textbf{Component}\\ \textbf{Transfer}}} 
        & Ref (Single Subject) & ``A \{subject\_type\} in \{scene\}, wearing \{clothing\_desc\}, with \{accessories\_desc\}." \\
        & Ref (Two Subjects) & ``A \{subject1\_type\} on the \{position1\} wearing \{cloth1\} with \{acc1\}, and a \{subject2\_type\} on the \{position2\} wearing \{cloth2\} with \{acc2\}, in \{scene\}." \\
        \cmidrule{2-3} 
        & Task (Complex Mode) & ``Extract \{elements\_desc\} from \{source\_desc\} in Image \{source\_label\}, then apply these elements to \{target\_desc\} in Image \{target\_label\}. Create a new composition showing the target subject(s) wearing/displaying the transferred elements." \\
        & Task (Simple Mode) & ``Task: Extract only the \{local\_element\} from the subject in Image A, then apply this element to the subject in Image B. Create a new composition showing the target subject wearing/displaying the \{local\_element\}." \\
        \midrule
        
        \multirow{2}{*}{\parbox{3cm}{\textbf{FG/BG}\\ \textbf{Composition}}} 
        & Reference Image & ``A photo of \{personalized\_obj\} in \{scene\}." \\
        \cmidrule{2-3} 
        & Task (2 Refs) & ``Generate an image where you cleanly extract the \{obj\_a\} from image A and replace the \{obj\_b\} in image B. The background from image B should remain unchanged." \\
        \midrule

        \multirow{2}{*}{\textbf{Story Inference}} 
        & Ref Image (4-panel) & ``Generate a four-panel comic with logically coherent and causally related events, following one of the specified types (physical property change, causal commonsense, or action continuation)." \\
        \cmidrule{2-3} 
        & Task (2/3 Refs) & ``Given the reference images, infer and generate a realistic photo of what might happen next." \\
        
        \bottomrule
    \end{tabular}
\end{table*}

\begin{table*}[t]
    \centering
    \small
    \caption{\textbf{Example of Evaluation Checkpoints.} Detailed checklist for Object Composition task.}
    \label{tab:eval_guidelines}
    \renewcommand{\arraystretch}{1.3} 
    \begin{tabular}{l l p{9.5cm}}
        \toprule
        \textbf{Dimension} & \textbf{ID} & \textbf{Evaluation Question} \\
        \midrule
        
        \multirow{3}{*}{\textbf{A. Instruction Following}} 
        & \texttt{A\_check\_1} \textbf{(H1)} & Does the image contain all specified objects as required by the instructions? \\
        & \texttt{A\_check\_2} & Are the relative arrangement and requested relations between objects correctly followed? \\
        & \texttt{A\_check\_3} & Are there no obviously extra or missing salient elements? \\
        \midrule
        
        \multirow{3}{*}{\textbf{B. Identity / Fidelity}} 
        & \texttt{B\_check\_1} \textbf{(H2)} & Does each object's identity strictly match its reference (e.g., category, instance)? \\
        & \texttt{B\_check\_2} & Are the key attributes (e.g., color, texture, shape) of each object well preserved? \\
        & \texttt{B\_check\_3} & Are the object details accurate and easily recognizable? \\
        \midrule
        
        \multirow{2}{*}{\textbf{C. Structure / Geometry}} 
        & \texttt{C\_check\_1} & Are the spatial relationships between objects consistent with the instructions and physically plausible? \\
        & \texttt{C\_check\_2} & Are the relative sizes, proportions, and perspective of objects realistic? \\
        \midrule
        
        \multirow{2}{*}{\parbox{3.5cm}{\textbf{D. Cross-Reference}\\ \textbf{Consistency}}} 
        & \texttt{D\_check\_1} & Are objects from different reference images integrated without conflicts or contradictions? \\
        & \texttt{D\_check\_2} & Are style, lighting, and background consistent across composed objects? \\
        \midrule
        
        \multirow{2}{*}{\textbf{G. Overall Usability}} 
        & \texttt{G\_check\_1} & Does the final scene appear natural, coherent, and visually plausible? \\
        & \texttt{G\_check\_2} & Are lighting, shadows, and global aesthetics of sufficient quality for practical use? \\
        \bottomrule
    \end{tabular}
\end{table*}

\begin{figure*}[t]
  \centering
  \includegraphics[width=0.86\textwidth]{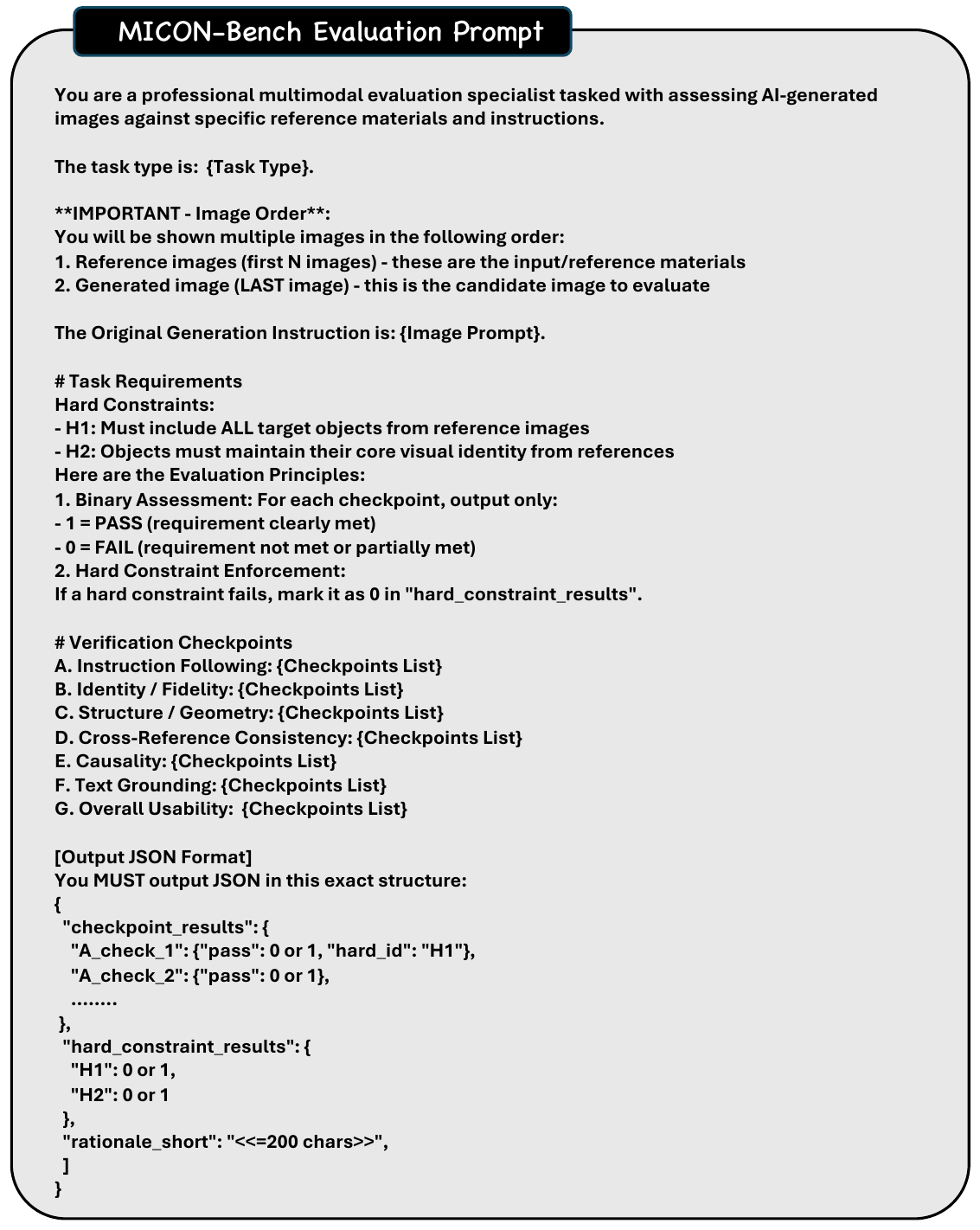}
  \caption{
The evaluation prompt used for MLLM scoring.
}
  \label{fig:sup1}
\end{figure*}

\end{document}